\documentclass[10pt,twocolumn,letterpaper]{article}

\usepackage{cvpr}              
\usepackage{bbm}
\usepackage[dvipsnames]{xcolor}

\usepackage{amssymb}
\usepackage{amsmath}
\usepackage{comment}
\usepackage{graphicx}
\usepackage{pifont}
\usepackage{float}

\usepackage{booktabs}
\usepackage{enumerate}
\usepackage{enumitem}
\usepackage{makecell}
\usepackage{tabu}
\usepackage{multirow}
\usepackage{utfsym}
\usepackage{colortbl}  
\definecolor{mygray}{gray}{.9}
\usepackage{pifont}
\usepackage{soul}

\usepackage[dvipsnames]{xcolor}

\usepackage{amssymb}
\usepackage{amsmath}
\usepackage{comment}
\usepackage{graphicx}
\usepackage{pifont}
\usepackage{float}

\usepackage{booktabs}
\usepackage{enumerate}
\usepackage{enumitem}
\usepackage{makecell}
\usepackage{tabu}
\usepackage{multirow}
\usepackage{utfsym}
\usepackage{colortbl}  
\definecolor{mygray}{gray}{.9}
\usepackage{pifont}
\usepackage{soul}

\definecolor{cvprblue}{rgb}{0.21,0.49,0.74}
\usepackage[pagebackref,breaklinks,colorlinks,citecolor=cvprblue]{hyperref}

\def\paperID{1417} 
\def\confName{CVPR}
\def\confYear{2024}

\title{SC-Tune: Unleashing Self-Consistent Referential Comprehension  in Large Vision Language Models}

\author{Tongtian Yue\textsuperscript{1,3\thanks{Equal Contribution.}} \quad Jie Cheng\textsuperscript{2,3*} \quad Longteng Guo\textsuperscript{1,3*} \quad Xingyuan Dai\textsuperscript{2,3} 
\\ Zijia Zhao\textsuperscript{1,3} \quad Xingjian He\textsuperscript{1,3} \quad Gang Xiong\textsuperscript{2,3} \quad Yisheng Lv\textsuperscript{2,3} \quad Jing Liu\textsuperscript{1,3\thanks{Corresponding author.}} \\
\textsuperscript{1}Laboratory of Cognition and Decision Intelligence for Complex Systems, CASIA \\
\textsuperscript{2}State Key Laboratory of Multimodal Artificial Intelligence Systems, CASIA
\\ \textsuperscript{3}School of Artificial Intelligence, University of Chinese Academy of Sciences}

\begin{document}


\maketitle
\begin{abstract}

Recent trends in Large Vision Language Models (LVLMs) research have been increasingly focusing on advancing beyond general image understanding towards more nuanced, object-level referential comprehension. In this paper, we present and delve into the self-consistency capability of LVLMs, a crucial aspect that reflects the models' ability to both generate informative captions for specific objects and subsequently utilize these captions to accurately re-identify the objects in a closed-loop process. This capability significantly mirrors the precision and reliability of fine-grained visual-language understanding.
Our findings reveal that the self-consistency level of existing LVLMs falls short of expectations, posing limitations on their practical applicability and potential. To address this gap, we introduce a novel fine-tuning paradigm named \textbf{Self-Consistency Tuning (SC-Tune)}. It features the synergistic learning of a cyclic describer-locator system. This paradigm is not only data-efficient but also exhibits generalizability across multiple LVLMs. 
Through extensive experiments, we demonstrate that SC-Tune significantly elevates performance across a spectrum of object-level vision-language benchmarks and maintains competitive or improved performance on image-level vision-language benchmarks. Both our model and code will be publicly available at \url{https://github.com/ivattyue/SC-Tune}.
\end{abstract}    
\section{Introduction}
\label{sec:intro}

\begin{figure}[htbp]
    \centering
    \includegraphics[width=0.5\textwidth]{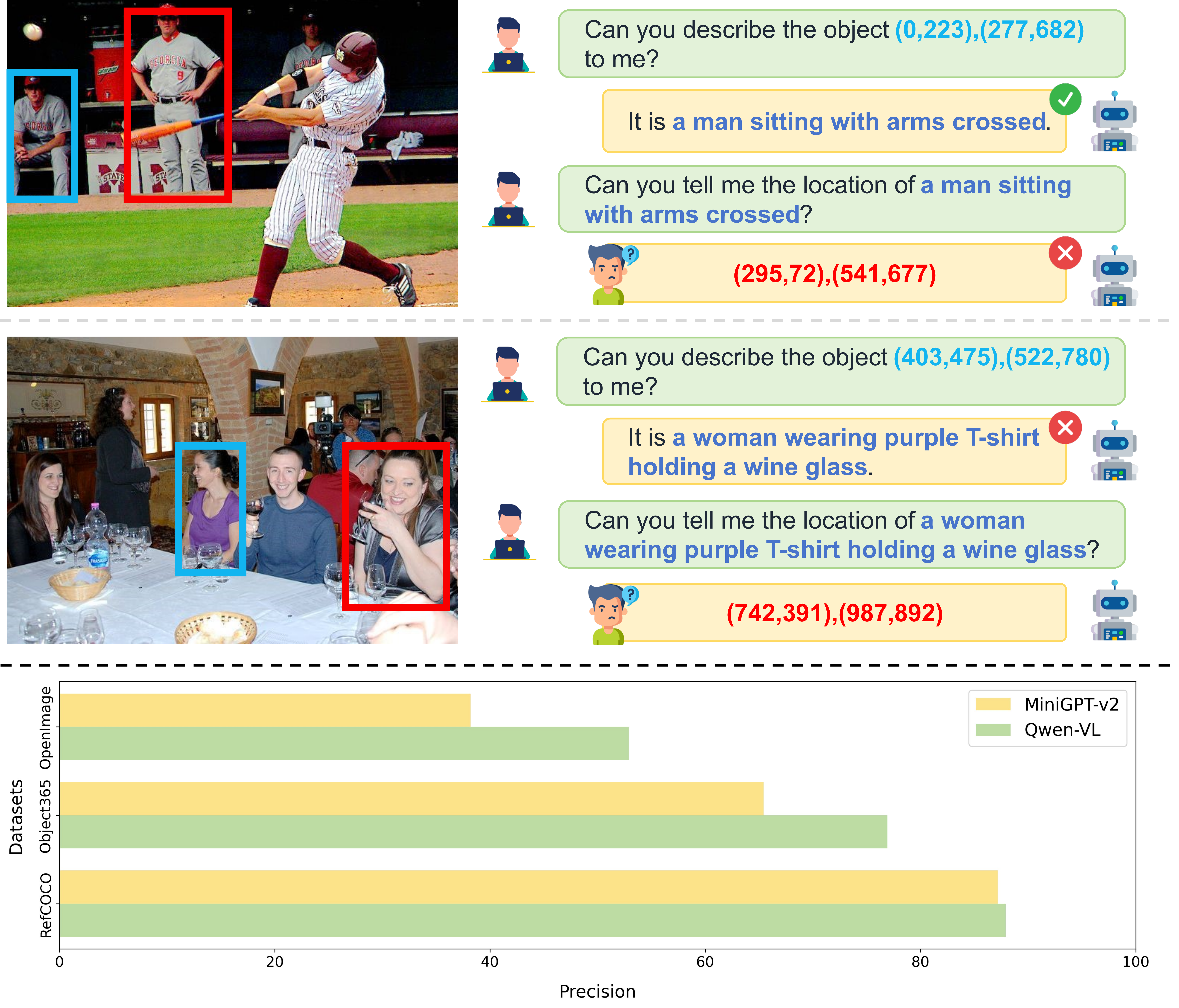}
    \caption{Upper: two examples demonstrating the shortfall in self-consistent referential comprehension of LVLMs, which are attributed to limited grounding (the first example) and limited captioning (the second example) capabilities, respectively. Lower: We use Pr@0.5 as the self-consistency evaluation metric. We consider it to be correct when the IoU between prediction bbox and the groung truth is greater than 0.5. The self-consistency levels of different LVLMs show
    pronounced performance gap between in-domain (RefCOCO) and out-of-domain (Object365 and OpenImages) datasets.
    }
    \label{fig:intro examples}
\end{figure}

Recently, Large Vision Language Models have witness remarkable
progress \cite{minigpt4, instructblip, llava}. 
By introducing learnable parameters to map visual features to the semantic space of Large Language Models (LLMs) \cite{gpt3, t5, llama, vicuna}, LVLMs demonstrate strong capabilities for visual content perception and expression. 
A line of works \cite{shikra, minigptv2, qwenvl, gpt4roi} go beyond basic image-level comprehension, and have emerged to cultivate object-level referential comprehension capability with LVLMs, namely \textit{understanding} or \textit{identifying} a specific object within an image. 
Referential comprehension capability bases on two symmetry object-level tasks, \textit{i.e.}, referring expression generation (REG) and referring expression comprehension (REC). REG is to describe an object in an image with discriminative referring expression, while REC is the reverse task of REG, aiming at localizing a particular object given an expression. 
Recent LVLMs handle spatial coordinate inputs and outputs in natural language, thereby unifying both REG and REC tasks in sequence modeling framework, which are optimized with multi-task instruction tuning on a large-scale collection of fine-grained region-text data \cite{refcoco, refcocog, visual_genome, kosmos-2}. 

A LVLM model that generates a referring expression for an object (REG) and then accurately locates back the object based on that expression (REC) demonstrates a deeper understanding of the content and context of the image. Such bbox-caption-bbox cyclical consistency is a crucial capability of LVLMs performing object-level understanding tasks, which we refer to as \textbf{\textit{self-consistency}} on referential comprehension (see Figure \ref{fig:intro examples}). It ensures that both the generation and comprehension aspects are aligned and accurate, leading to more robust models. If a model generates an incorrect referring expression or fails to accurately localize an object, the inconsistency becomes apparent when attempting to reverse the task. In real-world scenarios where accuracy in both generating and comprehending referential expressions is crucial (such as in autonomous navigation, visually impaired assistance, embodied agents, or interactive AI systems), self-consistency ensures reliability and usability of the model.

Surprisingly, through preliminary experiments, we found that the \textbf{\textit{self-consistency capability of current object-level LVLMs (\eg MiniGPT-v2 \cite{minigptv2} and Qwen-VL~\cite{qwenvl}) dramatically drops on out-of-domain images}}. As illustrated in Figure \ref{fig:intro examples}, we randomly select 4k bounding boxes from both the in-domain (seen during training) dataset RefCOCO \cite{refcoco} and the out-of-domain (unseen during training) datasets, \ie, OpenImages \cite{openimages} and Object365 \cite{objects365}, respectively. Staring from these bboxes, we sequentially perform REG and REC and examine whether the model can locate the correct region based on its generated region caption. As is shown in the figure, there exists pronounced performance gap of self-consistency between in-domain and out-of-domain images, implying poor generalization of their referential comprehension ability.

In this paper, we foster the self-consistent referential comprehension capability of object-level LVLMs by proposing a model fine-tuning paradigm named \textbf{\textit{Self-Consistency Tuning (SC-Tune)}}. We go beyond conventional multi-task learning of both REG and REC tasks, which neglects their consistent correlation. Self-consistency tuning centers on the synergistic learning of a cyclic dual-component system: a describer and a locator, which are two roles of the same pre-trained LVLM. In this system, the describer generates contextually enriched captions from bboxes. Subsequently, the locator operates on those generated captions, utilizing them as directives for precise object localization within images. 
The describer is trained through a bbox-caption-bbox self-consistency reward cycle, under Proximal Policy Optimization (PPO) \cite{ppo} reinforcement learning paradigm. This mechanism ensures the generation of captions that not only describe but also discriminatively guide the localization process. 
Conversely, the locator is supervisory trained to parse nuanced linguistic cues from the evolving synthetic captions of the describer, enabling precise and robust object localization even in out-of-domain images. 

The two components are updated under an iterative training cycle: freezing one component when training the other one. This alternating cycle is complemented by the synchronization of their parameters post each training cycle, fostering a harmonious growth of the overall system in fine-grained understanding.
By refining the interplay between caption generation and object localization, we enhance the model's self-consistency in both in-domain and out-of-domain images. 
Such Self-consistency tuning thereby increases the performance and robustness of LVLMs in fine-grained referential comprehension. 

 Our main contributions can be summarized as follows:

\setlist{nolistsep}
\begin{itemize}[noitemsep,leftmargin=*]
    \item 
    We propose self-consistency as a crucial metric for model reliability in 
    fine-grained referential comprehension and systematically examine this capability on existing LVLMs. 
    \item 
    We propose self-consistency tuning to effectively foster the self-consistent referential comprehension capability of LVLMs, which is data efficient and generalizable across multiple LVLMs.
    \item 
    By incorporating our proposed self-consistency tuning with state-of-the-art LVLMs, we observe notable enhancements in zero-shot performance across multiple  object-level vision-language benchmarks, while maintaining competitive or even improved performance on image-level vision-language benchmarks. 
\end{itemize}

\section{Related Work}
\label{relatedwork}

\paragraph{Referential Comprehension of LVLMs.}
In everyday human interactions, referencing specific objects or areas within a visual context is a frequent occurrence. Therefore, it is significant to augment LVLMs with robust referential comprehension capabilities. The mainstream paradigms for integrating location information into the understanding of LVLMs can be summarized as the followings, namely textual coordinate representation and regional feature extraction. The representative work of the former can be traced back to Pix2Seq \cite{pix2seq}. It leverages discrete coordinate tokens to encode spatial information, thereby unifying the referential comprehension training into the sequence modeling task. Notable works include OFA \cite{ofa}, Unified-io \cite{unifiedio}, Shikra \cite{shikra} and Kosmos-2 \cite{kosmos-2}. For the latter paradigm, represented by PVIT \cite{pvit} and GPT4RoI \cite{gpt4roi}, it utilizes Region of Interest (ROI) \cite{roi} to extract object-level features, then align these features to the textual modality. However, these works ignore the native self-consistency between reference generation and grounding when training referential comprehension capabilities. The disruption in the bidirectional consistency between visual and textual modalities indicates unsatisfactory alignment quality.

\paragraph{Reinforcement Learning in LLMs.}
Recent advancements in Large Language Models (LLMs) have been significantly influenced by Reinforcement Learning from Human Feedback (RLHF). This method has proven effective in aligning LLMs with human-centric values and preferences \cite{bai2022training, instructgpt}. 
RLHF hinges on training a reward model (RM) aligned with human preferences, subsequently refining LLMs based on the reward signals provided by the RM. Notably, LLMs are often updated using PPO algorithm \cite{stiennon2020learning,nakano2021webgpt}, with the addition of a Kullback–Leibler (KL) penalty to control deviations from the initial model \cite{instructgpt,liu2022second}. Extending beyond text, RLHF has been adapted for image generation. By introducing a RM, the alignment quality of prompts and images is improved \cite{aligning}, making it more consistent with human preferences \cite{imagereward}. To save the resources on collecting high-quality human preferences, \cite{lee2023rlaif} proposes to use AI instead of human in labeling preferences, a method termed RL from AI Feedback (RLAIF), which has demonstrated comparable efficacy. Similarly, in this work, we employ the locator to provide feedback as the reward signal for the describer. By treating the caption generation as a sequential decision-making problem, we leverage reinforcement learning to fine-tune the describer towards self-consistency.

\section{Self-Consistency Tuning }
\label{sec:method}

\begin{figure*}[htbp]
    \centering
    \includegraphics[width=0.86\textwidth]{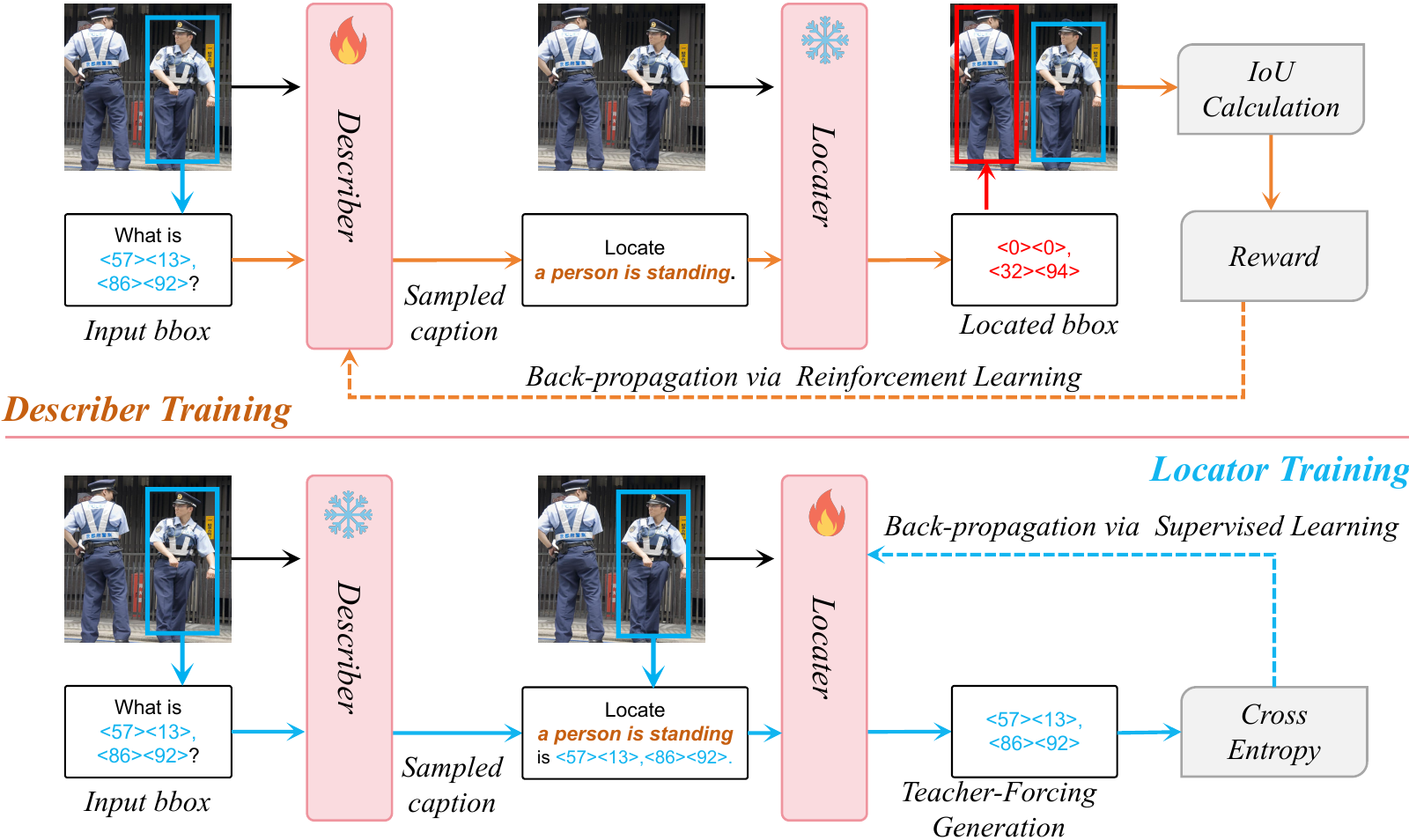}
    \caption{
    An illustration of our \textit{SC-Tune} framework. It mainly comprises describer training cycle and locator training cycle. Each cycle employs a bbox-caption-bbox pipeline with respective loss function designed for fostering the self-consistent referential comprehension capability of object-level LVLMs. Training alternates between these two cycles, with parameter synchronization post each training cycle.
    }
    \label{fig:model}
\end{figure*}

In this section, we introduce our training framework aimed at eliminating the sense of isolation between tasks, enhancing the model's self-consistent referring comprehension capability. This approach cyclically fine-tunes a dual-component system towards self-consistency. We briefly review the architectural design of current LVLMs in section \ref{subsec:LVLM Architecture}. Subsequently, in section \ref{subsec:training pipeline}, we provide a detailed introduction to our SC-Tune pipeline, as illustrated in Figure \ref{fig:model}. Then we elaborate the training process and loss function for each component respectively in section \ref{subsec:training of describer} and \ref{subsec:training of locator}.


\subsection{LVLM Architecture}
\label{subsec:LVLM Architecture}
The structure of current LVLMs can be primarily summarized into three parts: the visual encoder, LLM and the bridging component that connects the two modalities. Initially, input images are processed through the visual encoder for feature extraction. These visual features are then mapped onto the semantic space of LLM through the bridging component, where they are concatenated with the input text and fed into LLM for content generation. LVLMs acquire referential comprehension capabilities through region-text data, which is sourced from both small-scale human annotations \cite{refcoco, visual_genome, refcocog} and extensive weakly supervised data \cite{kosmos-2}. LVLMs proportionally resize the top-left and bottom-right coordinates to integers within a fixed range (\eg, [0,1000] for Qwen-VL and [0,100] for MiniGPT-v2). Subsequently these discrete and finite coordinates are represented in textual form. In REG and REC tasks, these textually represented coordinates serve as either input or expected output for LVLMs. Both tasks utilize a language modeling loss for optimization. However, the REC and REG capabilities trained in this manner are suboptimal due to a lack of symmetry. 


\begin{figure}[ht]
    \centering
    \includegraphics[width=0.45\textwidth]{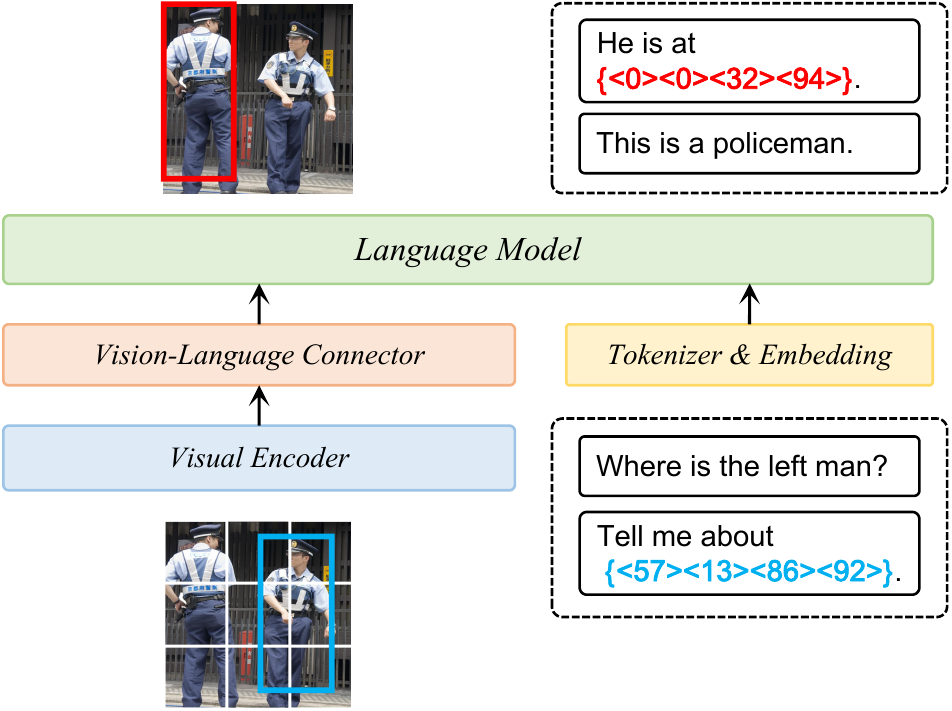}
    \caption{Architecture of object-level LVLMs. They mainly comprises three parts: the visual backbone for feature extraction, the vision-language connector for semantic alignment and a LLM. It has preliminary REC and REG capabilities by generating and understanding coordinates represented by text.}
    \label{fig:method LVLM}
\end{figure}

\subsection{Overview of \textbf{\textit{SC-Tune}}}
\label{subsec:training pipeline}
Our framework regards two roles, namely ``describer" and ``locator", of one pre-trained LVLM as a dual-component system. The describer focuses on generating detailed captions based on given bboxes, while the locator aims to reconstruct these bboxes from the self-generated captions. The central goal of our fine-tuning process is to find a set of parameters that has both enhanced capabilities.

To achieve that goal, we draw inspiration from the ``target network" utilized in Deep-Q Networks (DQN) \cite{DQN}. In DQN, one network is frozen as the ``target network" to provide training signals for the other network. As the other one inches closer to its goal, the frozen target network is updated by the parameters copied from the other network to facilitate continuous improvement. Similarly in our framework, we treat each component as the target network for the other to generate training signals. Moreover, we iteratively train the two components and synchronize their parameters when switching the focus of training. This method not only inherits the advantage of the target network to make training more stable, but also promotes the mutual improvement of the two capabilities within one model.

Specifically, the parameters of the describer and locator are the same initially. During the training stage of describer, we freeze the locator and only tune the describer for a pre-defined number of steps, $H$. Following this, we synchronize the parameter of describer to the locator. Conversely, in the training stage of locator, we freeze the describer and only tune the locator for $H$ steps. After this stage, we synchronize the parameter of locator to the describer. In each training stage, we employ a bbox-caption-bbox pipeline (show in Figure \ref{fig:model}), leveraging the task symmetry and consistent correlation for self-consistency fine-tuning. This cyclical training process continues until it effectively integrates the improvements of both capabilities into a cohesive model.

\subsection{Describer Training}
\label{subsec:training of describer}

This subsection details the training process for the describer, which is responsible for generating detailed captions $\hat{C}$ from given bounding boxes $B$. Concurrently, the locator is frozen to localize bboxes $\hat{B}$ for these self-generated captions $\hat{C}$. In this reconstruction process, our aim is to maximize the similarity between the input and output bboxes, \ie, $\text{IoU}(B,\hat{B})$, for self-consistency, which implicitly requires these captions $\hat{C}$ to be informative and discriminative.

However, the operation of sampling from vocabulary during generation is non-differentiable, resulting that the describer cannot be updated directly through back-propagating reconstruction loss. To address this challenge, we employ reinforcement learning (RL), which does not require derivation of the action sampling process. 

\paragraph{PPO Loss.}
In the setting of RL, both the describer and locator are regarded as ``agents" with policy $\pi$ parameterized by $\theta$ and $\phi$, respectively. Initially, $\theta=\phi$. At timestep $t$, the previously generated tokens $\{\hat{c}_1,\dots,\hat{c}_{t-1}\}$, the input image $I$, and the bbox $B$ are regarded as the ``state" $s_t$. The generating caption token $\hat{c}_t$ is served as the ``action" taken with the policy $\pi_\theta$, \textit{i.e.}, $\hat{c}_t\sim \pi_\theta(\cdot|s_t)$. The objective is to maximize the cumulative rewards $\sum_{t=1}^T R(\hat{c}_t,s_t)$, where $T$ is the terminal token generation step. To achieve the objective, we use PPO algorithm to optimize the policy in the trust region for stable training. The RL loss function is formulated as follows:
\begin{equation}
\resizebox{0.9\hsize}{!}{
$
    \mathcal{L}^{RL}(\theta) = -\mathbb{E}_t \left[ 
    \min \left(
    r_t(\theta) A_t, \ 
    \text{clip}\left( r_t(\theta), 1-\epsilon, 1+\epsilon \right) A_t
    \right)
    \right]
$}
\label{eq:ppo loss}
\end{equation}
where $r_t(\theta)$ is the importance sampling ratio: $r_t(\theta)=\frac{\pi_\theta(\hat{c}_t|s_t)}{\pi_{\theta_\text{old}}(\hat{c}_t|s_t)}$, and $\theta_\text{old}$ is the policy parameters before update. $\epsilon$ is a hyperparameter and $A_t$ is the advantage function. clip function is to limit values to a range.

\paragraph{Reward.}
Given the self-generated caption $\hat{C} = \{ \hat{c}_1, \dots, \hat{c}_T \}$ and input image $I$, the frozen locator $\pi_\phi$ generates a predicted bbox $\hat{B}$, \textit{i.e.}, $\hat{B}=f(\pi_\phi(\hat{C},I))$, where $f$ represents regular expression search from generated sentence to parse bbox. Due to the goal of reconstructing the original bbox $B$, it is intuitive to set the $\text{IoU}(B,\hat{B})$ as the reward function. Notably, it is a sentence-level reward for describer. However, this reward function does not constrain the semantic information of captions, probably causing forgetfulness in linguistic rule. Therefore, we add a token-level KL penalty from the initial model to mitigate over-optimization of the reconstruction reward. The complete reward function is formulated as follows:
\begin{align}
    R(\hat{c}_t,s_t) = \mathbbm{1}_{t=T}(t)\cdot\text{IoU}(B,\hat{B}) - \beta\log\frac{\pi_\theta(\hat{c}_t|s_t)}{\pi_{\theta_\text{ref}}(\hat{c}_t|s_t)}
\end{align}
where $\mathbbm{1}$ is the indicator function, $\theta_\text{ref}$ is the initial parameters, and $\beta$ is the KL coefficient.

\paragraph{Value-Free Advantage.}
The advantage function is used to evaluate the long-term benefits of a chosen action compared to other actions. In standard PPO algorithm, it needs another value function to compute the advantage, which requires additional network and computing resources. Alternatively, we apply the form of advantage function in ``REINFORCE with baseline" algorithm \cite{sutton2018reinforcement} formulated in Eq. (\ref{eq:advantage}), where the baseline is employed to reduce variance theoretically.
\begin{align}
    A_t(\hat{c}_t,s_t) = \sum_{i=t}^T R(\hat{c}_i,s_i) - R(c^*_T,s^*_T)
    \label{eq:advantage}
\end{align}
where the superscript * represents the baseline. We choose greedily decoded captions $C^*$ generated from initial model $\theta_\text{ref}$ as the baseline. Note that there is no need of gradients for action $\hat{c}_t$, reward $R(\hat{c}_t,s_t)$, advantage $A(\hat{c}_t,s_t)$, and old sampling probability $\pi_{\theta_\text{old}}(\hat{c}_t|s_t)$, only the latest sampling probability $\pi_\theta(\hat{c}_t|s_t)$ needs to calculate gradient in the training loss (\ref{eq:ppo loss}).

\subsection{Locator Training}
\label{subsec:training of locator}
In this subsection, we detail the training process for the locator. Unlike the describer that is in the middle of the bbox-caption-bbox pipeline, the gradient of the loss function can be propagated back to the locator directly. Therefore, we adopt self-supervised learning on locator for simplicity.

Specifically, during the training of locator $\pi_\phi$, we freeze the describer $\pi_\theta$ to generate pseudo captions $\hat{C}$ from given bboxes $B$ and image $I$. Then we fine-tune locator with the data $(B, I, \hat{C})$ using loss derived from maximum likelihood estimation (MLE). Assuming the tokens of tokenized instructions building upon $(B, I, \hat{C})$ are denoted as $\{w_1,\dots,w_T\}$, the training loss of locator is formulated as follows:
\begin{align}
    \mathcal{L}^\text{MLE}(\phi) = -\mathbb{E}_t\left[ \log P(w_t|w_1,\dots,w_{t-1}) \right]
    \label{eq:LM loss}
\end{align}

Our framework iteratively fine-tunes the describer and locator for $H$ steps using PPO loss (\ref{eq:ppo loss}) and MLE loss (\ref{eq:LM loss}), respectively. Post each training phase, we synchronize their parameters to facilitate synergistic and continuous improvement, as detailed in section \ref{subsec:training pipeline}.
\section{Experiments}

\subsection{Experimental Settings}
\paragraph{Baseline LVLMs.} We verify the effectiveness and universality of our proposed training framework on two prevalent fine-grained LVLM, \ie Qwen-VL \cite{qwenvl} and MiniGPT-v2 \cite{minigptv2}. Qwen-VL adopts Qwen-7B \cite{qwen} as its foundation component, and uses the Vision Transformer (ViT) \cite{vit} as the visual encoder, with pre-trained weights from Openclip’s ViT-bigG \cite{openclip}. For MiniGPT-v2, it takes the visual tokens from EVA-Clip \cite{eva} and leverage LLaMA2-Chat (7B) \cite{llama2} as LLM. Uniformly, for both models, we inherit the checkpoints from stage 3 and execute \textit{SC-Fune} on this basis to further refine the referential comprehension capabilities. 

\paragraph{Implementation Details.} We adopt AdamW \cite{adamw} as the optimizer. Parameter synchronization is performed every 200 steps and the training stage is then switched. We train both baseline models for 1 epoch which contains 6 stage switches. Both models are trained with a batch size of 128 and weight decay of 0.1. The learning rate is set to 5e-7 and 1e-6 for describer and locator training stage, respectively. We set the KL coefficient and PPO epoch to 0.01 and 2 for reinforcement learning. The input image is resized to 448 × 448 without any additional data-augmentation. The model is trained on 8 NVIDIA A100 GPUs, lasting roughly 15 hours for Qwen-VL and 10 hours for MiniGPT-v2.

\paragraph{Training Dataset.} 
Our \textit{SC-Tune} is data-efficient. We leverage about 166K images and corresponding bbox locations sampled from Object365 \cite{objects365} Dataset. Training does not require any text annotations. 

\paragraph{Evaluation Datasets and Metrics.}
We evaluate the baseline models on various object-level and image-level benchmarks to validate the referential comprehension enhanced by our framework. For REC and REG, we consider RefCOCO/+/g \cite{refcoco, refcocog}, ReferItGame \cite{referitgame} and Flickr30K Entities \cite{flickr}. For referential question answering, we evaluate on Visual-7W \cite{visual7w} and PointQA-Local \cite{looktwice}. Furthermore, we conduct assessments on the image-level benchmarks to substantiate the enhanced fine-grained alignment quality brought by our \textit{SC-Tune}. For Image Caption, we employ Nocaps \cite{nocaps} and Flickr30K \cite{flickr}. Regarding the question-answering tasks, we adopt VQAv2 \cite{vqav2} and GQA \cite{gqa} as the evaluation benchmarks. 

For all benchmarks, we follow the prompt templates used in the original instruction tuning of the baseline models. Greedy search is used for decoding. In REC task, the bounding box predicted by the model is considered as correct for reporting accuracy if its intersection over union (IoU) between prediction and ground-truth is higher than 0.5. In REG and Image Caption tasks, following previous works, we report results using METEOR \cite{meteor} and CIDEr \cite{cider} metrics. In QA tasks, we leverage top-1 accuracy to measure the matching degree bewteen the response and ground-truth.

\subsection{Self-Consistency Evaluation Results}

We first investigate the consistency enhancement brought by \textit{SC-Tune}, as detailed in Table \ref{tab:sc}. By default, we perform subsequent performance evaluations using checkpoints obtained from training with Object365. We additionally present the results of Qwen-VL trained on OpenImages as supplement. More detailed experiments can be referred to in the Appendix. 
For both baseline models, the application of \textit{SC-Tune} on a small scale Object365 dataset results in a notable improvement in self-consistency. The enhancement is applicable to both in-domain data RefCOCO, and out-of-domain data OpenImages. \textit{SC-Tune} demonstrates robustness to data distribution, which is reflected in the similar average increases in self-consistency level:  13 points for Object365 and 12.5 points for OpenImages in Qwen-VL.

\begin{table}[t]
    \small
    \setlength{\belowcaptionskip}{1.0pt}
    \begin{center}
    \caption{Self-consistency evaluation on various benchmarks. We report self-consistency level using accuracy, where a sample is considered as right when IoU between prediction and ground-truth is higher than 0.5.}
    \vspace{-5pt}
    \setlength{\tabcolsep}{0.8mm}{
    \begin{tabular}{l|c|c|c}
    \specialrule{.1em}{.05em}{.05em}
        Method & Object365 & OpenImages & RefCOCO  \\
        \specialrule{.1em}{.05em}{.05em}
        Qwen-VL & 76.9 & 52.9 & 87.9 \\
        \quad +\textit{SC-Tune} (Object365) & \textbf{94.1} & 68.8  & \textbf{93.8} \\
        \quad +\textit{SC-Tune} (OpenImages) & 89.6 & \textbf{73.6} & 92.0    \\
        \hline
       MiniGPT-v2 & 65.4 & 38.2 & 87.2 \\
        \quad +\textit{SC-Tune} (Object365) & \textbf{76.6} & \textbf{48.5} & \textbf{90.4} \\
        \specialrule{.1em}{.05em}{.05em}
    \end{tabular}
    \label{tab:sc}}
    \end{center}
    \vspace{-20pt}
\end{table}\

\subsection{Object-Level Evaluation Results}
\paragraph{Reference Expression Comprehension Results}
Table \ref{tab:rec_ood} demonstrates the improvement of referring grounding ability of baseline models under the out-of-domain setting before and after \textit{SC-Tune}. For Qwen-VL, compared to the baseline, using \textit{SC-Tune} achieves an accuracy improvement of 7.8\% and 13.34\% on the test split of two prevalent datasets, \ie ReferItGame and Flickr30K, respectively. It is worth noting that under the zero-shot setting, the performance of Qwen-VL equipped with our \textit{SC-Tune} can even be comparable with the fully supervised model represented by CLIP-VG \cite{clipvg}. Similar patterns are also shown on another baseline MiniGPT-v2, with performance improvements of 7.6\% and 4.6\% respectively. These results provide direct evidence that \textit{SC-Tune} can refine a more general referential comprehension capability. 

\vspace{-2mm}
\begin{table}[H]
    \small
    \setlength{\belowcaptionskip}{1.0pt}
    \begin{center}
    \caption{REC results on ReferItGame and Flickr30k Entities Datasets. We report the accuracy metric for all methods.}
    \vspace{-5pt}
    \setlength{\tabcolsep}{1.6mm}{
    \begin{tabular}{l|c|c|c}
    \specialrule{.1em}{.05em}{.05em}
        Method & Zero-Shot & ReferIt & Flickr30K  \\
        \specialrule{.1em}{.05em}{.05em}
        \multicolumn{4}{l}{\color{gray} \textit{Specialists}} \\
        \hline 
        SeqTR ~\cite{seqtr}& \ding{55}  & 69.66  & 81.23 \\
        VLTVG ~\cite{VLTVG}  & \ding{55}   & 71.98 & 79.84  \\
        CLIP-VG ~\cite{clipvg} & \ding{55}   & 70.89 & 81.99 \\
        \hline
        \multicolumn{4}{l}{\color{gray} \textit{Generalists}} \\
        \hline
        Qwen-VL & \ding{51} & 61.48  & 64.15  \\
        \rowcolor{mygray} Qwen-VL + \textit{SC-Tune} & \ding{51} & \textbf{69.28} &  \textbf{77.49}  \\
        \hline
        MiniGPT-v2 & \ding{51} & 36.05 & 55.39    \\
        \rowcolor{mygray} MiniGPT-v2 + \textit{SC-Tune} & \ding{51} &  \textbf{43.68} & \textbf{60.03}  \\
        \specialrule{.1em}{.05em}{.05em}
    \end{tabular}
    \label{tab:rec_ood}}
    \end{center}
    \vspace{-20pt}
\end{table}\

Although the main focus of our training framework is not to achieve further improvement on a specific benchmark by adapting to its distribution, from the perspective of completeness, we report the results on the in-domain REC dataset, \eg RefCOCO \cite{refcoco}, in Table \ref{tab:rec_id}. Introducing \textit{SC-Tune} has a slight performance drop compared to the baselines on the three evaluation subsets of RefCOCO. Intuitively, as tuning progresses, the caption style generated by the model gradually deviates from the in-domain style. The adaptation to this deviation will inevitably lead to a decrease of in-domain capability. Therefore, we argue that this fluctuation is acceptable as a trade-off with the gain in generalization ability. Furthermore, following RLHF pipeline which mixes the pretraining gradients into the original gradients to fix the performance regressions on benchmarks \cite{rlhf}, we implement this regularization in our training as an attempt. Specifically, during the locator training stage, we augmented the synthesized captions with additional supervised data from RefCOCO. As illustrated in the Table \ref{tab:rec_id}, this strategy enables further enhancement of in-domain performance. However, as previously mentioned, this is not the primary focus of our research.
\vspace{-2mm}
\begin{table}[htbp]
    \small
    \setlength{\belowcaptionskip}{1.0pt}
    \begin{center}
    \caption{REC results on RefCOCO Dataset under in-domain setting. $\mathcal{L}_{sup}$ represents the incorporation of in-domain RefCOCO data in the locator training stage. The performance metrics for the two baseline models are reproduced using their official open-source code and checkpoints.}
    \vspace{-5pt}
    \setlength{\tabcolsep}{2mm}{
    \begin{tabular}{l|ccc}
    \specialrule{.1em}{.05em}{.05em}
        \multirow{2}{*}{Method} & \multicolumn{3}{c}{RefCOCO} \\
        \cline{2-4}
        ~ & val & test A & test B \\
        \specialrule{.1em}{.05em}{.05em}
        \multicolumn{4}{l}{\color{gray} \textit{Specialists}} \\
        \hline
        G-DINO-L ~\cite{groundingdino} & 90.56 & 93.19 & 88.24   \\
        UNINEXT-H ~\cite{uninext} & 92.64 & 94.33 & 91.46   \\
        ONE-PEACE ~\cite{onepeace} & 92.58 & 94.18 & 89.26   \\
        \hline
        \multicolumn{4}{l}{\color{gray} \textit{Generalists}} \\
        \hline 
        VisionLLM-H ~\cite{visionllm} & - & 86.70 & -   \\
        OFA-L ~\cite{ofa} & 79.96 & 83.67 & 76.39   \\
        Shikra-7B ~\cite{shikra} & 87.01 & 90.61 & 80.24   \\
        \hline
        Qwen-VL & 88.88 & 92.27 & 84.30 \\
        \rowcolor{mygray} Qwen-VL + \textit{SC-Tune} & 88.04 & 90.77 & 84.62  \\
        Qwen-VL + \textit{SC-Tune} + $\mathcal{L}_{sup}$ & 89.61 & 93.43 & 85.72  \\
        \hline
        MiniGPT-v2 & 84.84 & 89.49 & 81.08     \\
        \rowcolor{mygray} MiniGPT-v2 + \textit{SC-Tune} & 83.59 & 88.50 & 80.22    \\
        MiniGPT-v2 + \textit{SC-Tune} + $\mathcal{L}_{sup}$ & 85.29 & 90.77 & 82.33    \\
        \specialrule{.1em}{.05em}{.05em}
    \end{tabular}
    \label{tab:rec_id}}
    \end{center}
    \vspace{-20pt}
\end{table}

\paragraph{Reference Expression Generation Results.}
Table \ref{tab:reg_ood} evaluates our \textit{SC-Tune} on REG benchmarks under the out-of-domain setting. Symmetrically, we adopt ReferItGame and Flickr30K for evaluation. Baseline model shows strong grounded captioning capabilities after \textit{SC-Tune}, performing better than other zero-shot counterparts. For Qwen-VL, compared with the baseline, our \textit{SC-Tune} outperforms it by over 20 CIDEr points across all tasks of ReferItGame and Flickr30K. Similarly, significant improvements can also be observed on MiniGPT-v2, \ie an increase of 9 points in average CIDEr. 

\begin{table*}[htbp]
    \small
    \setlength{\belowcaptionskip}{1.0pt}
    \centering
    \caption{REG results on ReferItGame and Flickr30k Entities Datasets under out-of-domain setting. Following previous work, ReferItGame test split is divided into test A and test B based on whether the sample is human. }
    \vspace{-5pt}
    \setlength{\tabcolsep}{1.116mm}{
    \begin{tabular}{l|cc|cc|cc|cc}
    \toprule
        \multirow{3}{*}{Method}  & \multicolumn{4}{c|}{ReferIt} & \multicolumn{4}{c}{Flickr30K}  \\
        \cline{2-9}
        ~ & \multicolumn{2}{c|}{test A} & \multicolumn{2}{c|}{test B} & \multicolumn{2}{c|}{val} & \multicolumn{2}{c}{test}  \\
        \cline{2-9}
        ~ & Meteor & CIDEr & Meteor & CIDEr & Meteor & CIDEr & Meteor &  CIDEr \\
        \hline
        \multicolumn{4}{l}{\color{gray} \textit{Specialists}} \\
        \hline
        SLR ~\cite{slr} & 3.5 & 10.9 & 2.8  & 8.5 & 3.0 & 14.1 & 2.7 & 13.9 \\
        EU ~\cite{refgta}  & 2.0 & 9.5 & 1.9  & 7.7 & 2.7  & 14.8 & 2.4  & 15.0 \\
        DisCLIP ~\cite{disclip}  & 9.7 & 8.8 & 9.0  & 6.3 & 9.5  & 6.7 & 9.6  & 6.8 \\
        \hline
        \multicolumn{4}{l}{\color{gray} 
        \textit{Generalists}} \\
        \hline
        Qwen-VL &  7.5 & 19.2 & 9.3 & 35.0 & 16.8  & 53.6 &  17.6  & 57.6 \\
        
        \rowcolor{mygray} Qwen-VL + \textit{SC-Tune} &  \textbf{11.6} & \textbf{43.2} & \textbf{13.1} & \textbf{51.7} &  \textbf{18.5}  &\textbf{ 73.7}  & \textbf{18.4} & \textbf{80.9} \\
        \hline
         MiniGPT-v2 & 8.9 & 33.3 & 8.2 & 47.5 & 15.8 & 78.7 & 16.3 & 84.0  \\
        \rowcolor{mygray} MiniGPT-v2 + \textit{SC-Tune} & \textbf{11.1} & \textbf{39.5 }& \textbf{11.2} & \textbf{56.8} & \textbf{16.3} & \textbf{87.1} & \textbf{16.9} &\textbf{ 93.5} \\
    \bottomrule
    \end{tabular}
    \label{tab:reg_ood}}
    \vspace{-10pt}
\end{table*}

We report the performance on the RefCOCO dataset in Table \ref{tab:reg_id} as a complementary evaluation of the in-domain REG capability. Facilitating more informative and unique descriptions by \textit{SC-Tune}, we found that the in-domain captioning capabilities of baseline models can be further improved. For MiniGPT-v2, this benefit makes it a generalist model that can even compete with specialist models represented by PFOS \cite{PFOS} in terms of CIDEr scores, \eg, 87.7 for PFOS vs 82.5 for MiniGPT-v2 in TestA split and 132.9 for PFOS vs 131.1 for MiniGPT-v2 in TestB split. The improvement in Qwen-VL is also worthy of recognition.

\begin{table}[H]
    \small
    \setlength{\belowcaptionskip}{1.0pt}
    \begin{center}
    \caption{REG results on RefCOCO Dataset under in-domain setting.}
    \vspace{-5pt}
    \setlength{\tabcolsep}{1.6mm}{
    \begin{tabular}{l|cc|cc}
    \specialrule{.1em}{.05em}{.05em}
        \multirow{3}{*}{Method} & \multicolumn{4}{c}{RefCOCO}  \\
        \cline{2-5}
        ~  &  \multicolumn{2}{c|}{TestA} & \multicolumn{2}{c}{TestB} \\
        \cline{2-5}
        ~ &  Meteor & CIDEr & Meteor & CIDEr \\
        \specialrule{.1em}{.05em}{.05em}
        \multicolumn{4}{l}{\color{gray} \textit{Specialists}} \\
        \hline
        SLR ~\cite{slr}  & 26.8 & 69.7 & 32.9 & 132.3  \\
        EU ~\cite{refgta} & 31.1 & 83.7 & 33.0 & 133.3  \\
        PFOS ~\cite{PFOS} & 30.3 & 87.7 & 34.1 & 132.9  \\
        \hline
        \multicolumn{4}{l}{\color{gray} \textit{Generalists}} \\
        \hline
        Qwen-VL  & 23.8 & 88.0 & 24.9 & 110.0 \\
        \rowcolor{mygray} Qwen-VL + \textit{SC-Tune} & \textbf{26.3} & \textbf{104.0} & \textbf{28.7} & \textbf{129.1}  \\
        \hline
        MiniGPT-v2 & 16.9 & 62.1 & 21.4 & 113.7   \\
        \rowcolor{mygray} MiniGPT-v2 + \textit{SC-Tune} & \textbf{23.2} & \textbf{82.5} &\textbf{ 26.2 }&  \textbf{131.1}\\
        \specialrule{.1em}{.05em}{.05em}
    \end{tabular}
    \label{tab:reg_id}}
    \end{center}
    \vspace{-20pt}
\end{table}

\paragraph{Referential Question Answering Results.}
Apart from the two abilities of referential comprehension, \ie REC and REG, we also evaluate the ability of referential question answering. For Visual-7W \cite{visual7w}, it features a which setting, requiring the model to select one matching box from four options based on the given reference. For Local-QA \cite{looktwice}, the models are asked to answer questions based on the given bbox. We report the performance on both benchmarks in Table \ref{tab:qa_ood}. Through tuning Qwen-VL, the accuracy on Visual-7W and Local-QA increases by 11.3\% and 8.1\% respectively. The patterns shown on MiniGPT-v2 are consistent, which fully proves the universality of \textit{SC-Tune}.

\begin{table}[H]
    \small
    \setlength{\belowcaptionskip}{1.6pt}
    \begin{center}
    \caption{Referential QA Results on Visual-7W and Local-QA. The evaluation is under zero-shot setting. We report the accuracy metric for both benchmarks.}
    \vspace{-5pt}
    \setlength{\tabcolsep}{4mm}{
    \begin{tabular}{l|c|c}
    \specialrule{.1em}{.05em}{.05em}
        Method & Visual-7W & Local \\
        \hline
        Qwen-VL  & 34.77 & 60.75  \\
        \rowcolor{mygray} Qwen-VL + \textit{SC-Tune} & \textbf{46.09} & \textbf{72.13} \\
        \hline
        MiniGPT-v2 & 27.53 & 49.63  \\
        \rowcolor{mygray} MiniGPT-v2 + \textit{SC-Tune} & \textbf{35.29} & \textbf{51.15} \\
        \specialrule{.1em}{.05em}{.05em}
    \end{tabular}
    \label{tab:qa_ood}}
    \end{center}
    \vspace{-20pt}
\end{table}

\subsection{Image-Level Evaluation Results}
Intuitively, when the quality of referential comprehension is optimized, it enhances the perceptual and understanding capabilities at the image level. Consequently, we conducted extensive evaluation on multiple image-level benchmarks, with the results presented in Table \ref{tab:image_id}. The results indicate that \textit{SC-Tune} demonstrates improvements across multiple coarse-grained benchmarks. Given that SC-Tune requires only 166K images and does not necessitate text annotations, this enhancement is satisfactory.

\begin{table}[H]
    \small
    \setlength{\belowcaptionskip}{1.0pt}
    \begin{center}
    \caption{Image-level evaluation results on image caption and QA task. For QA tasks, accuracy metrics are employed, while for image caption, we use CIDEr score. For VQA and GQA, following previous works, we report the metrics on the val split and testdev split, respectively.}
    \vspace{-5pt}
    \setlength{\tabcolsep}{1mm}{
    \begin{tabular}{l|c|c|c|c}
    \specialrule{.1em}{.05em}{.05em}
        Method  & Nocaps & Flick30K & VQA & GQA \\
        \hline
        \textit{Specialists SOTAs} & \textcolor{gray}{127.0}  & \textcolor{gray}{84.5}  & \textcolor{gray}{86.1}  & \textcolor{gray}{72.1}  \\
        \hline
        \multicolumn{4}{l}{\color{gray} \textit{Generalists}} \\
        \hline
        BLIP-2& 103.9 & 71.6 & 65.0 & 32.3 \\
        InstructBLIP  & 121.9 & 82.8 & - & 49.5  \\
        Shikra & - & 73.9 & 77.4 & -  \\
        \hline
        Qwen-VL & 120.2 & 81.0 & 78.2 & 57.5  \\
        \rowcolor{mygray} Qwen-VL + \textit{SC-Tune} &  \textbf{121.4} & \textbf{86.0} & \textbf{79.0} & \textbf{58.4}  \\
        \hline
        MiniGPT-v2 & 93.5 & 77.1 & 72.6 & 59.1  \\
        \rowcolor{mygray} MiniGPT-v2 + \textit{SC-Tune}  & \textbf{94.6} & \textbf{78.4}  & \textbf{73.4} &  \textbf{59.9}  \\
        \specialrule{.1em}{.05em}{.05em}
    \end{tabular}
    \label{tab:image_id}}
    \end{center}
    \vspace{-20pt}
\end{table}

\subsection{Ablation Studies}

\paragraph{Synergistic Effect of Iterative Training.}
In this section, we conduct an ablation study on the strategy of iterative training with parameter exchange, to demonstrate the synergistic effect of our \textit{SC-Tune}. The results are presented in Table \ref{tab:ab_it} based on Qwen-VL. Taking the second row as an example, we freeze the locator while tuning the describer, training it for an equivalent number of steps as in the iterative approach. This process resulted in a specialized model with enhanced captioning capabilities, and vice versa. The results, however, indicate that even specialized model for captioning does not perform as well as the model with \textit{SC-Tune} on REG tasks. It is equally applicable to the grounding specialized model. These results demonstrate the superiority of consistently improve both capabilities over the isolated optimization of either.

\begin{table}[H]
    \small
    \setlength{\belowcaptionskip}{1.0pt}
    \begin{center}
    \caption{Ablation study on synergistic effect of iterative training. We report the performance in REC, REG of ReferItGame and Local-QA based on Qwen-VL. For REC and QA tasks, accuracy metrics are employed, while for REG, we use CIDEr score.}
    \vspace{-5pt}
    \setlength{\tabcolsep}{1mm}{
    \begin{tabular}{cc|c|cc|c}
    \specialrule{.1em}{.05em}{.05em}
        \multicolumn{2}{c|}{Tuning Component}  & REC@ReferIt & \multicolumn{2}{c|}{REG@ReferIt} & QA@Local \\
        \cline{1-6}
        \centering
        Locator & Describer & test & test A & test B & test \\
        \hline
         \ding{55} & \ding{55} & 61.48 & 19.2 & 35.0 & 60.75 \\
         \ding{55} & \ding{51} & 52.92 & 38.2 & 45.8 & 70.12 \\
         \ding{51} & \ding{55} & 64.82 & 19.8 & 35.1 & 69.54 \\
        \rowcolor{mygray} \ding{51} & \ding{51} & \textbf{69.28} & \textbf{43.2} & \textbf{51.7} & \textbf{72.13} \\
        \specialrule{.1em}{.05em}{.05em}
    \end{tabular}
    \label{tab:ab_it}}
    \end{center}
    \vspace{-20pt}
\end{table}

\paragraph{Training Steps for Each Cycle.}
In this section, we analyze the training steps for each cycle. The analysis pertaining to the training steps is presented in Table \ref{tab:ab_step}. The results indicate that selecting appropriate training steps to balance fluctuations in various capabilities is of significance. 

\begin{table}[H]
    \small
    \setlength{\belowcaptionskip}{1.0pt}
    \begin{center}
    \caption{Ablation study on training steps based on Qwen-VL.}
    \vspace{-5pt}
    \setlength{\tabcolsep}{2mm}{
    \begin{tabular}{c|c|cc|c}
    \specialrule{.1em}{.05em}{.05em}
        \multirow{2}{*}{Steps}  & REC@ReferIt & \multicolumn{2}{c|}{REG@ReferIt} & QA@Local \\
        \cline{2-5}
        & test & test A & test B & test \\
        \hline
        50 & 68.41 & 34.3 & 46.6 & 71.15 \\
        \rowcolor{mygray} 200 & \textbf{69.28} & \textbf{43.2} & \textbf{51.7} & \textbf{72.13}\\
        1000 & 67.45 & 39.3 & 48.3 & 71.12 \\
        \specialrule{.1em}{.05em}{.05em}
    \end{tabular}
    \label{tab:ab_step}}
    \end{center}
    \vspace{-20pt}
\end{table}

\paragraph{Training Data Source.} 
In this section, we conduct an ablation study on the source of training data based on Qwen-VL. It demonstrates the universality of \textit{SC-Tune} across various data distributions. Specifically, we selected another out-of-domain object detection dataset OpenImages \cite{openimages}, and an in-domain dataset Visual Genome \cite{visual_genome}. In terms of data filter and training strategies, we keep consistent with which we conduct in Object365. The experiment results in Table \ref{tab:ab_data} indicate that \textit{SC-Tune} achieves a significant performance improvement over the baseline model across all three data distributions.

\begin{table}[H]
    \small
    \setlength{\belowcaptionskip}{1.0pt}
    \begin{center}
    \caption{Ablation study on data source based on Qwen-VL.}
    \vspace{-5pt}
    \setlength{\tabcolsep}{1.2mm}{
    \begin{tabular}{l|c|cc|c}
    \specialrule{.1em}{.05em}{.05em}
        \multirow{2}{*}{Data Source} & REC@ReferIt & \multicolumn{2}{c|}{REG@ReferIt} & QA@Local \\
        \cline{2-5}
        & test & val & test & test \\
        \hline
         & 61.48 & 19.2 & 35.0 & 60.75 \\
         \hline
        \rowcolor{mygray} Object365
        &  \textbf{69.28} & \textbf{43.2} & \textbf{51.7} & 72.13 \\
        OpenImages
        &  66.07 & 35.5 & 47.6 & \textbf{72.50}\\
        Visual Genome
        &  63.28 & 26.3 & 41.2 & 68.56 \\
        \specialrule{.1em}{.05em}{.05em}
    \end{tabular}
    \label{tab:ab_data}}
    \end{center}
    \vspace{-20pt}
\end{table}

\label{sec:method}
\section{Conclusion}

In this work, we reveal a notable shortfall in the self-consistency levels of current LVLMs. Responding to this gap, we propose the Self-Consistency Tuning (SC-Tune), an object-level fine-tuning paradigm designed to improve self-consistent referential comprehension capability. Central to SC-Tune is a cyclic training loop of a dual-component system (\ie describer and locator), which promotes a harmonious growth of the overall system in fine-grained understanding. SC-Tune not only is data-efficient but also demonstrates robust generalizability across multiple LVLMs. Our comprehensive experiments show that SC-Tune significantly improves performance across various object-level vision-language benchmarks, while simultaneously sustaining or augmenting performance in image-level vision-language benchmarks. We plan to release our model and code to the public, expecting to foster future research in this direction.

\textbf{Acknowledgements}\quad We thank all the insightful reviewers for the helpful suggestions.
This work was supported by the National Science and Technology Major Project (No.2022ZD0118801), National Natural Science Foundation of China (U21B2043, 62206279).
\clearpage
{
    \small
    \bibliographystyle{ieeenat_fullname}
    \bibliography{main}
}
\clearpage
\setcounter{page}{1}

\section*{Overview}
In this supplementary material, we provide following items:
\begin{itemize}[noitemsep,leftmargin=*]
    \item (Sec.\textcolor{red}{1}) Details about the preliminary experiment.
    \item (Sec.\textcolor{red}{2}) More results of self-consistency enhancement.
    \item (Sec.\textcolor{red}{3}) Visualization of the \textit{SC-Tune} effects.
    \item (Sec.\textcolor{red}{4}) Implementation details of training data selection. 
\end{itemize}

\section{Preliminary Experiment Details}

In this section, we delineate the details of our preliminary experiment. Initially, we randomly select 4K bounding boxes (bbox) from three datasets (\ie RefCOCO \cite{refcoco}, OpenImage \cite{openimages} and Object365 \cite{objects365}) to respectively build the test splits, ensuring that each bbox originates from different images. For any given baseline model, we fill the image along with the coordinates of the bbox into its referring expression generation (REG) instruction template, guiding the model to generate a region caption for the specified bbox. Subsequently, we fill the image and the generated caption into its referring expression comprehension (REC) instruction template, enabling it to locate the coordinates described by the caption. Following previous works \cite{kosmos-2, uninext}, we compute the intersection over union (IoU) between newly predicted coordinates and the original ones. If the IoU exceeds 0.5, the model is considered to have achieved the required level of self-consistency for the given sample. Ultimately, we employ the proportion of samples meeting this criterion within the test split (also known as $Pr@0.5$) as the metric for measuring the self-consistency level of a model.

\section{Self-Consistency Enhancement}

In this section, we present the full results across baseline models \cite{qwenvl, minigptv2} and data distributions \cite{openimages, objects365,refcoco} to evaluate the self-consistency enhancement brought by \textit{SC-Tune}. For the two baseline models, both OpenImage and Object365 serve as out-of-domain data sources, whereas RefCOCO remains visible throughout their training process. We conduct data filtering (detailed in Sec \ref{filtering}) on OpenImage and Object365, aligning their data sizes to be comparable with that of RefCOCO for fairness. It is worth to emphasize that prior to the filtering process, we have already excluded the 4k
test samples utilized in the preliminary experiment. Ultimately, we obtain about 166K training samples from Object365 and 138K training samples from OpenImage, respectively. These samples, in conjunction with the training split of RefCOCO, are utilized for the self-consistency enhancement evaluation shown in Table \ref{tab:sce}.

\begin{table}[H]
    \small
    \setlength{\belowcaptionskip}{1.0pt}
    \begin{center}
    \caption{Self-consistency evaluation on three benchmarks. We report self-consistency level using accuracy, where a sample is considered as right when IoU between prediction and ground-truth is higher than 0.5.}
    \vspace{-5pt}
    \setlength{\tabcolsep}{0.8mm}{
    \begin{tabular}{l|c|c|c}
    \specialrule{.1em}{.05em}{.05em}
        Method & Object365 & OpenImages & RefCOCO  \\
        \specialrule{.1em}{.05em}{.05em}
        Qwen-VL & 76.9 & 52.9 & 87.9 \\
        \quad +\textit{SC-Tune} (Object365) & \textbf{94.1} & 68.8  & \textbf{93.8} \\
        \quad +\textit{SC-Tune} (OpenImages) & 89.6 & \textbf{73.6} & 92.0    \\
        \quad +\textit{SC-Tune} (RefCOCO) & 83.4 & 58.8 &   93.5 \\
        \hline
       MiniGPT-v2 & 65.4 & 38.2 & 87.2 \\
        \quad +\textit{SC-Tune} (Object365) & \textbf{76.6} & 48.5 & 90.4 \\
        \quad +\textit{SC-Tune} (OpenImages) & 74.9 & \textbf{50.2} & 91.1 \\
        \quad +\textit{SC-Tune} (RefCOCO) & 69.8 & 45.5 & \textbf{91.6} \\
        \specialrule{.1em}{.05em}{.05em}
    \end{tabular}
    \label{tab:sce}}
    \end{center}
    \vspace{-20pt}
\end{table}

It clearly indicates that employing \textit{SC-Tune} enhances the self-consistency levels for both baseline models. Taking Qwen-VL \cite{qwenvl} as an example, tuning with the out-of-domain datasets (\ie Object365 and OpenImage) results in a self-consistency level increase of over 18\% in respective test splits. Moreover, applying \textit{SC-Tune} on in-domain data RefCOCO similarly elevates self-consistency levels across three test splits. The results sufficiently demonstrate the robustness and applicability of \textit{SC-Tune} across different data sources. Additionally, a similar pattern is observable in MiniGPT-v2, further confirming the compatibility of \textit{SC-Tune} to various models.

\section{Visualization}

\begin{figure}[ht]
    \centering
    \includegraphics[width=0.45\textwidth]{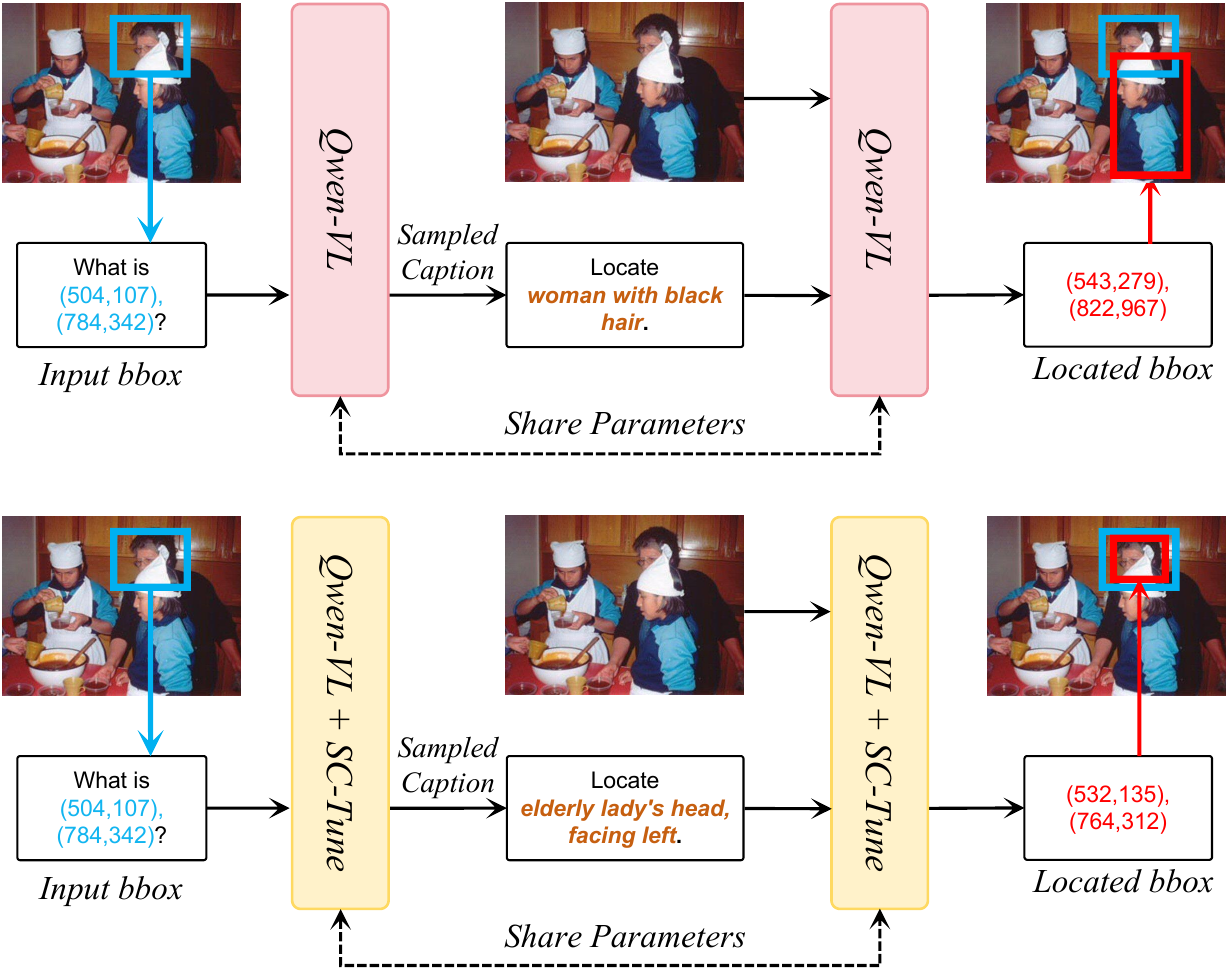}
    \vspace{2mm} 
    \rule{\linewidth}{1pt} 
    \vspace{2mm}
    \includegraphics[width=0.45\textwidth]{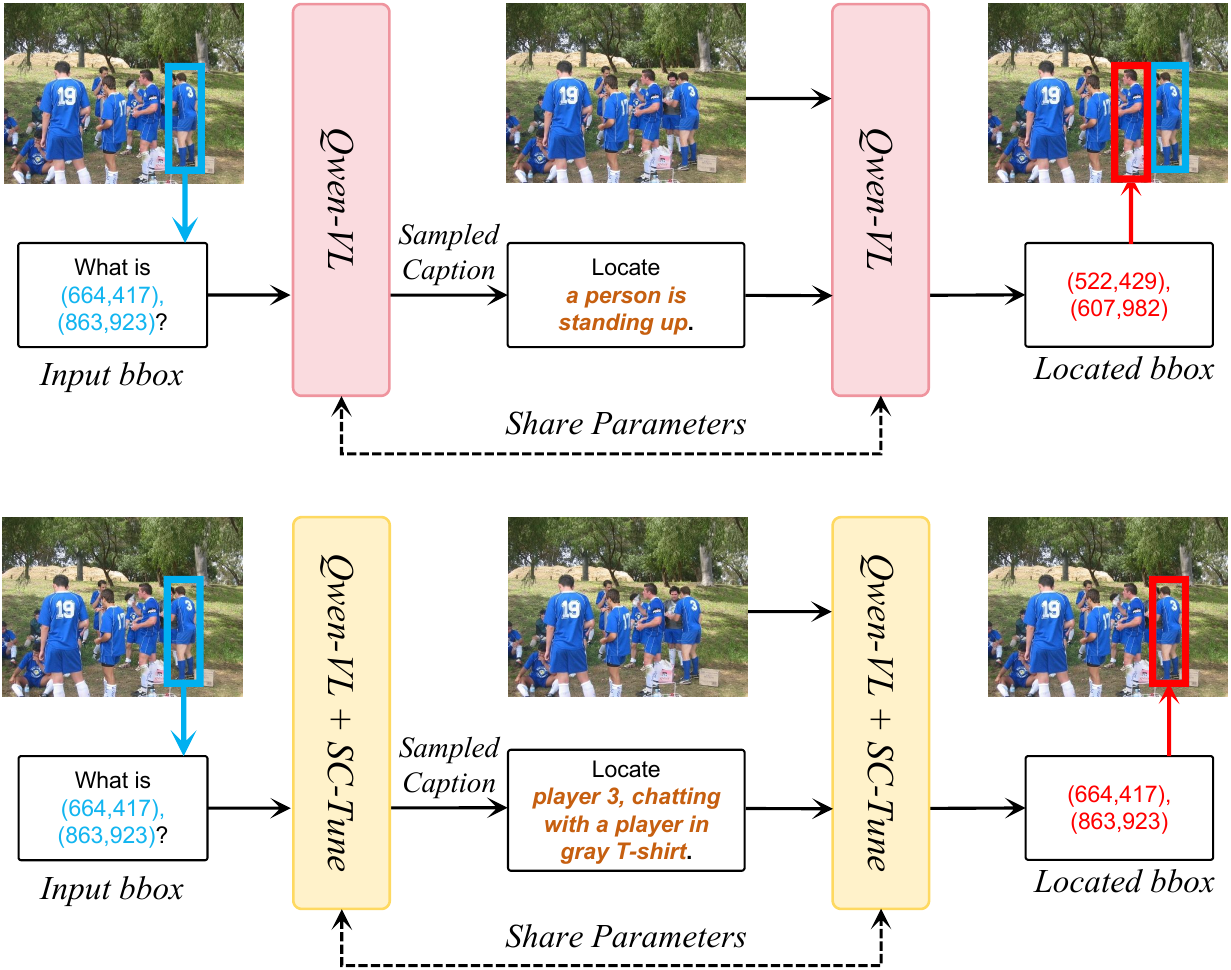}
    \caption{Case study to demonstrate the enhanced REG capability after \textit{SC-Tune}.}
    \label{fig:supply reg}
\end{figure}

In this section, we leverage Qwen-VL as an example to demonstrate the improvements of REG and REC capabilities after the application of \textbf{SC-Tune}. Figure \ref{fig:supply reg} qualitatively showcases the enhanced REG capability, which can be summarized in two aspects: 

(1) A refined understanding of detailed and unique attributes. Specifically, as illustrated in the upper part of Figure \ref{fig:supply reg}, the original model generates a generic description, which is applicable to all three women depicted in the given image without distinction. However, after \textit{SC-Tune}, the model captures more detailed information about the selected object, such as age, body parts, and facial orientation for unique identification.

(2) An enhanced capability to integrate visual context. For instance, in the lower part of Figure \ref{fig:supply reg}, compared with the description generated by original model, the \textit{SC-Tuned} model incorporates additional contextual information. It includes sufficient contextual clues like the jersey number of the player and his interaction with surrounding individuals.

\begin{figure}[ht]
    \centering
    \includegraphics[width=0.45\textwidth]{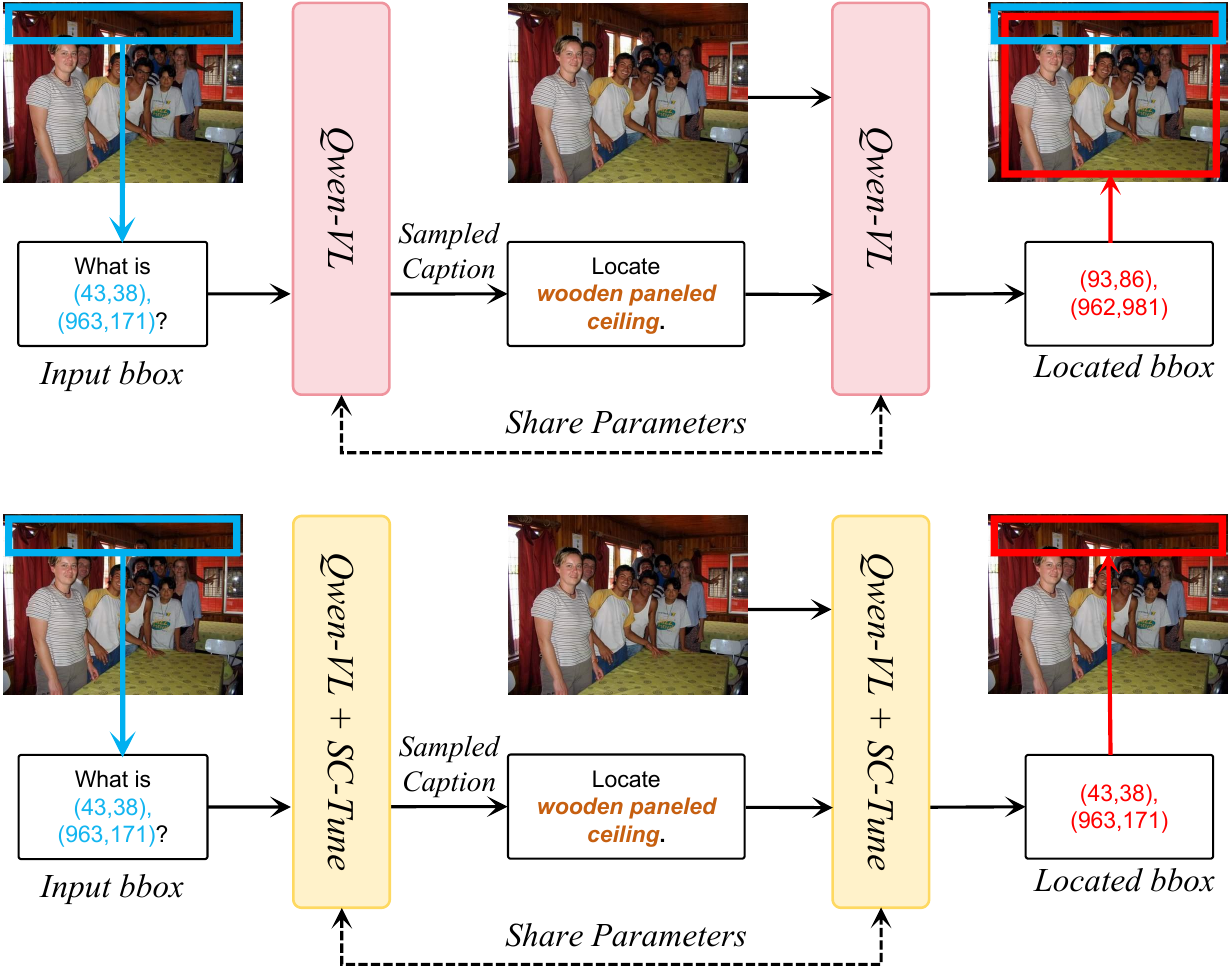}
    \vspace{2mm} 
    \rule{\linewidth}{1pt} 
    \vspace{2mm} 
    \includegraphics[width=0.45\textwidth]{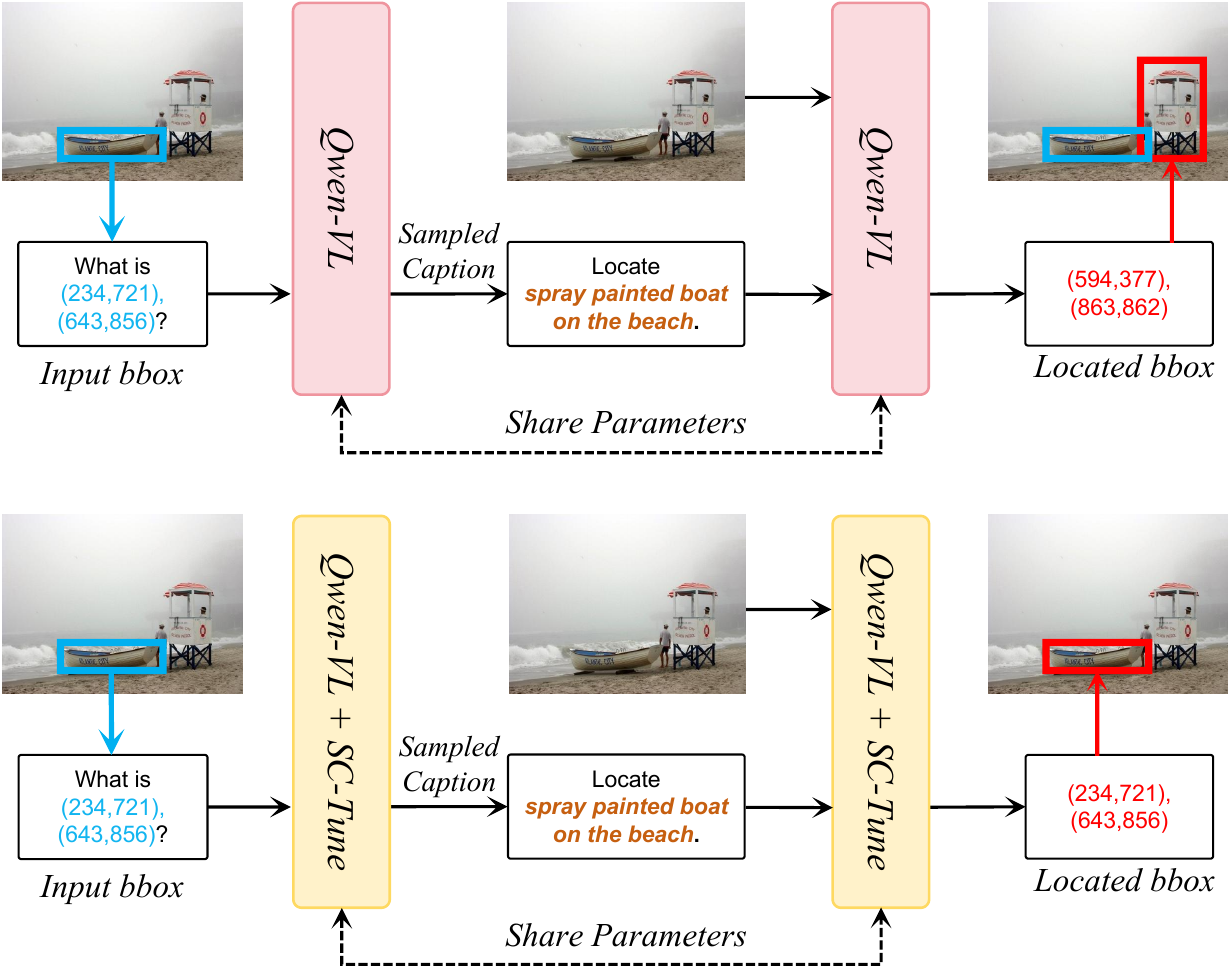}
    \caption{Case study to demonstrate the enhanced REC capability after \textit{SC-Tune}.}
    \label{fig:supply rec}
\end{figure}

Figure \ref{fig:supply rec} qualitatively showcases the enhanced REC capability. The cases illustrated in the figure demonstrates that for a given bbox and image, even if two models, before and after \textit{SC-Tune}, generate informative and equal descriptions, the model without \textit{SC-Tune} still fails to accurately locate back the original bbox. It is largely attributes to the lack of self-consistency in the multi-task learning paradigm, which hinders the effective alignment of these two capabilities \ie REC and REG. \textit{SC-Tune} significantly amends this mismatch, thereby enhancing the visual grounding ability.

\section{Training Data Filtering}

In this section, we introduce the process of data filtering. Specifically, for each image, if there are no categories that appear more than once, the image and all the corresponding annotations are discarded. Otherwise, we record the information as a triplet consisting of the image, category, and all bboxes that fit the category. This design is intended to better introduce ambiguity and increase the difficulty of training. Additionally, for the bboxes within an image, if two bboxes have an intersection over union (IoU) greater than 0.5, the triplets associated with these overlapping bboxes are removed. It is to address the potential confusion caused by overlap, which could interfere with model training. Lastly, we eliminate the triplet which contain a bbox that occupy less than 2\% of the image area. The same filtering approach is applied to both the Object365 and OpenImage datasets.
\label{filtering}

\end{document}